\documentclass{article}

\usepackage[preprint,nonatbib]{nips_2018}

\usepackage[utf8]{inputenc} 
\usepackage[T1]{fontenc}    
\usepackage{hyperref}       
\usepackage{url}            
\usepackage{booktabs}       
\usepackage{amsfonts}       
\usepackage{microtype}      
\usepackage{subfigure}
\usepackage{epsfig}
\usepackage{graphicx}
\usepackage{amsmath}
\usepackage{amssymb}
\usepackage{subfigure}
\usepackage{float}

\title{NetScore: Towards Universal Metrics for Large-scale Performance Analysis of Deep Neural Networks for Practical On-Device Edge Usage}

\author{
  Alexander Wong$^{1,2}$\\
  $^{1}$Waterloo Artificial Intelligence Institute, University of Waterloo, Waterloo, ON, Canada\\
  $^{2}$DarwinAI Corp., Waterloo, ON, Canada\\
  \texttt{a28wong@uwaterloo.ca} \\
}

\begin{document}

\maketitle

\begin{abstract}
Much of the focus in the design of deep neural networks has been on improving accuracy, leading to more powerful yet highly complex network architectures that are difficult to deploy in practical scenarios, particularly on edge devices such as mobile and other consumer devices given their high computational and memory requirements.  As a result, there has been a recent interest in the design of quantitative metrics for evaluating deep neural networks that accounts for more than just model accuracy as the sole indicator of network performance.  In this study, we continue the conversation towards universal metrics for evaluating the performance of deep neural networks for practical on-device edge usage.  In particular, we propose a new balanced metric called \textbf{NetScore}, which is designed specifically to provide a quantitative assessment of the balance between accuracy, computational complexity, and network architecture complexity of a deep neural network, which is important for on-device edge operation.  In what is one of the largest comparative analysis between deep neural networks in literature, the NetScore metric, the top-1 accuracy metric, and the popular information density metric were compared across a diverse set of 60 different deep convolutional neural networks for image classification on the ImageNet Large Scale Visual Recognition Challenge (ILSVRC 2012) dataset.  The evaluation results across these three metrics for this diverse set of networks are presented in this study to act as a reference guide for practitioners in the field.  The proposed NetScore metric, along with the other tested metrics, are by no means perfect, but the hope is to push the conversation towards better universal metrics for evaluating deep neural networks for use in practical on-device edge scenarios to help guide practitioners in model design for such scenarios.
\end{abstract}

\section{Introduction}

There has been a recent urge in both research and industrial interests in deep learning~\cite{lecun2015deep}, with deep neural networks demonstrating state-of-the-art performance in recent years across a wide variety of applications.  In particular, deep convolutional neural networks~\cite{lecun1,lecun2} has been shown to outperform other machine learning approaches for visual perception tasks ranging from image classification~\cite{krizhevsky2012imagenet} to object detection~\cite{liu2016ssd} and segmentation~\cite{MaskRCNN}.  One of the key driving factors behind the tremendous recent successes in deep neural networks has been the availability of massive computing resources thanks to the advances and proliferation of cloud computing and highly parallel computing hardware such as graphics processing units (GPUs).  The availability of this wealth of computing resources has enabled researchers to explore significantly more complex and increasingly deeper neural networks that has resulted in significant performance gains over past machine learning methods.  For example, in the realm of visual perception, the depth of deep convolutional neural networks with state-of-the-art accuracies have reached hundreds of layers, hundreds of millions of parameters in size, and billions of calculations for inferencing.

While the ability to build such large and complex deep neural networks has led to a constant increase in accuracy, the primary metric for performance widely leveraged for evaluating networks, it has also created significant barriers to the deployment of such networks for practical edge device usage.  The practical deployment bottlenecks associated with the powerful yet highly complex deep neural networks in research literature has become even more visible in recent years due to the incredible proliferation of mobile devices, consumer devices, and other edge devices and the increasing demand for machine learning applications in such devices.  As a result, the design of deep neural networks that account for more than just accuracy as the sole indicator of network performance and instead strike a strong balance between accuracy and complexity has very recently become a very hot area of research focus, with a number of different deep neural network architectures designed specifically with efficiency in mind~\cite{iandola2016squeezenet,howard2017mobilenets,tinyssd,munet,sandler2018mobilenetv2,squishednet,ShuffleNet}.

One of the key challenges in designing deep neural networks that strikes a strong balance between accuracy and complexity for practical usage lies in the difficulties with assessing how well a particular network architecture is striking that balance.  As previous mentioned, using accuracy as the sole metric for network performance does not provide the proper indicators of how efficient a particular network is in practical scenarios such as deployment on mobile devices and other consumer devices.  As a result, there has been a recent interest in the design of quantitative metrics for evaluating deep neural networks that accounts for more than just model accuracy.  In particular, it is generally desirable to design such metrics in a manner that is as hardware vendor agnostic as possible so that different network architectures can be compared to each other in a consistent manner.  One of the most widely cited metrics in research literature for assessing the performance of deep neural networks that accounts for both accuracy and architectural complexity is the information density metric proposed by~\cite{Canziani}, which attempts to measure the relative amount of accuracy captured within one of the most basic building blocks of a deep neural network: a parameter.  More specifically, the information density ($D(\mathcal{N})$) of a deep neural network $\mathcal{N}$ is defined as the accuracy of the deep neural network (denoted by $a(\mathcal{N})$) divided by the number of parameters needed for representing it (denoted by $p(\mathcal{N})$),
		\vspace{-0.1cm }
		\begin{equation}
D(\mathcal{N}) = \frac{a(\mathcal{N})}{p(\mathcal{N})}
\vspace{-0.1cm }
		\end{equation}

\noindent While highly effective for giving a good general idea of the balance between accuracy and architectural complexity (which also acts as a good indicator for memory requirements), the information density metric does not account for the fact that, depending on the design of the network architecture, the architecture complexity does not necessarily reflect the computational requirements for performing network inference (e.g., MobileNet~\cite{howard2017mobilenets} has more parameters than SqueezeNet~\cite{iandola2016squeezenet} but has lower computational requirements for network inference).  Therefore, the exploration and investigation towards universal  performance metrics that account for accuracy, architectural complexity, and computational complexity is highly desired as it has the potential to improve network model search and design.

In this study, we continue the conversation towards universal metrics for evaluating the performance of deep neural networks for practical usage.  In particular, we propose a new balanced metric called \textbf{NetScore}, which is designed specifically to provide a quantitative assessment of the balance between accuracy, computational complexity, and network architecture complexity of a deep neural network.  This paper is organized as follows.  Section 2 describes the proposed NetScore metric and the design principles around it.  Section 3 presents and discusses experimental results that compare the NetScore, information density, and top-1 accuracy across 60 different deep convolutional neural networks for image classification on the ImageNet Large Scale Visual Recognition Challenge (ILSVRC 2012) dataset~\cite{ImageNet}, making this one of the largest comparative studies between deep neural networks.
\newpage
\vspace{-0.2cm }
\section{NetScore: Design Principles}
\vspace{-0.1 cm}

The proposed NetScore metric (denoted here as $\Omega$) for assessing the performance of a deep neural network $\mathcal{N}$ for practical usage can be defined as:
		\vspace{-0.2cm }
		\begin{equation}
		\Omega(\mathcal{N}) = 20\log\left(\frac{{a(\mathcal{N})}^{\alpha}}{{p(\mathcal{N})}^{\beta}{m(\mathcal{N})}^{\gamma}}\right)
		\end{equation}
\vspace{-0.2cm }

where $a(\mathcal{N})$ is the accuracy of the network, $p(\mathcal{N})$ is the number of parameters in the network, $m(\mathcal{N})$ is the number of multiply–accumulate (MAC) operations performed during network inference, and $\alpha$, $\beta$, $\gamma$ are coefficients that control the influence of accuracy, architectural complexity, and computational complexity of the network on $\Omega$.  A number of design principles were taken into consideration in the design of the proposed NetScore metric, which is described below.

\textbf{Model accuracy representation}:  In the NetScore metric, the obvious incorporation of the model accuracy $a(\mathcal{N})$ of the network $\mathcal{N}$ into the metric is in the numerator of the ratio, as an increase in accuracy should naturally lead to an increase in the metric, similar to the information density metric~\cite{Canziani}.  We further introduce a coefficient $\alpha$ in the proposed NetScore metric to provide better control over the influence of model accuracy on the overall metric.  In particular, we set $\alpha=2$ to better emphasize the importance of model accuracy in assessing the overall performance of a network in practical usage, as deep convolutional neural networks that have unreasonably low model accuracy remain unusable in practical scenarios, regardless how small or fast the network is.  In this study, the unit used for $a(\mathcal{N})$ is in percent top-1 accuracy on the ILSVRC 2012 dataset~\cite{ImageNet}.

\textbf{Model architectural and computational complexity representations}:  Taking inspiration from the information density metric~\cite{Canziani}, we represent the architectural complexity of a deep neural network by the number of parameters $p(\mathcal{N})$ in the network $\mathcal{N}$ and incorporate it in the denominator of the ratio.  As such, the architecture complexity of the network is inversely proportional to the metric $\Omega$, where an increase in architectural complexity results in a decrease in $\Omega$.  In addition, we incorporate the computational complexity of the deep neural network as an additional factor in the denominator of the ratio to be taken into consideration for assessing the overall performance of a network for practical usage, which is particularly important in operational scenarios such as inference on mobile devices and other consumer devices where computational power is limited.  To represent the computational complexity of the network $\mathcal{N}$ in a manner that is relatively hardware vendor agnostic, thus enabling a more consistent comparison between networks, we chose to leverage the number of multiply–accumulate (MAC) operations necessary for performing network inference.  Given that the computational bottleneck associated with performing network inference on a deep neural network is predominantly in the computation of MAC operations, the number of MAC operations $m(\mathcal{N})$ is a good proxy for the computational complexity of the network.  By incorporating both architectural and computational complexity, the proposed NetScore metric can better quantify the balance between accuracy, memory requirements, and computational requirements in practical usage.  Furthermore, we introduce two coefficients ($\beta$ and $\gamma$, respectively) to provide better control over the influence of architectural and computational complexity on the overall metric.  In particular, we set $\beta=0.5$ and $\gamma=0.5$ since, while architectural and computational complexity are both very important factors to assessing the overall performance of a network in practical scenarios, the most important metric remains the model accuracy given that, as eluded to before, networks with unreasonably low model accuracy are not useful in practical scenarios regardless of size and speed.

\textbf{Logarithmic scaling}:
One of the difficulties with comparing the overall performance of different deep neural networks with each other is their great diversity in their model accuracy, architectural complexity, and computational complexity.  This makes the dynamic range of the performance metric quite large and unwieldy for practitioners to compare for model search and design purposes.  To account for this large dynamic range, we take inspiration from the field of signal processing; in particular, the decibel scale commonly used to express the ratio between one value of a property to another on a logarithmic scale.  In the proposed NetScore metric, we transform the ratio between the model accuracy property ($a(\mathcal{N})$) and the model architectural and computational complexity ($p(\mathcal{N})$ and $m(\mathcal{N})$) into the logarithmic decibel scale to reduce the dynamic range to within a more readily interpretable range.

\vspace{-0.2 cm}
\section{Experimental Results and Discussion}
\vspace{-0.1 cm}
To get a better sense regarding the overall performance of the huge wealth of deep convolutional neural networks introduced in research literature in the context of practical usage, we perform a large-scale comparative analysis across a diverse set of 60 different deep convolutional neural networks designed for image classification using the following quantitative performance metrics: i) top-1 accuracy, ii) information density, and iii) the proposed NetScore metric.  The dataset of choice for the comparative analysis in this study is the ImageNet Large Scale Visual Recognition Challenge (ILSVRC 2012) dataset~\cite{ImageNet}, which consists of 1000 different classes.  To the best of the author's knowledge, this comparative analysis is one of the largest in research literature and the hope is that the results presented in this study can act as a reference guide for practitioners in the field.

The set of deep convolutional neural networks being evaluated in this study are: AlexNet~\cite{krizhevsky2012imagenet}, AmoebaNet-A (4, 50)~\cite{AmoebaNet}, AmoebaNet-A (6, 190)~\cite{AmoebaNet}, AmoebaNet-A (6, 204)~\cite{AmoebaNet}, AmoebaNet-B (3, 62)~\cite{AmoebaNet}, AmoebaNet-B (6, 190)~\cite{AmoebaNet}, AmoebaNet-C (4, 50)~\cite{AmoebaNet}, AmoebaNet-C (6, 228)~\cite{AmoebaNet}, CondenseNet (G=C=4)~\cite{CondenseNet}, CondenseNet (G=C=8)~\cite{CondenseNet}, DenseNet-121 (k=32)~\cite{densenet}, DenseNet-169 (k=32)~\cite{densenet}, DenseNet-161 (k=48)~\cite{densenet}, DenseNet-201 (k=32)~\cite{densenet}, DPN-131~\cite{DPN}, GoogleNet~\cite{GoogleNet}, IGC-L100M2~\cite{IGC}, IGC-L16M16~\cite{IGC}, IGC-L100M2~\cite{IGC}, Inception-ResNetv2~\cite{Inceptionv4}, Inceptionv2~\cite{Inceptionv2}, Inceptionv3~\cite{Inceptionv2}, Inceptionv4~\cite{Inceptionv4}, MobileNetv1 (1.0-224)~\cite{howard2017mobilenets}, MobileNetv1 (1.0-192)~\cite{howard2017mobilenets}, MobileNetv1 (1.0-160)~\cite{howard2017mobilenets}, MobileNetv1 (1.0-128)~\cite{howard2017mobilenets}, MobileNetv1 (0.75-224)~\cite{howard2017mobilenets}, MobileNetv2~\cite{sandler2018mobilenetv2}, MobileNetv2 (1.4)~\cite{sandler2018mobilenetv2}, NASNet-A (4 @ 1056)~\cite{NASNet}, NASNet-A (6 @ 4132)~\cite{NASNet}, NASNet-B (4 @ 1536)~\cite{NASNet}, NiN~\cite{NiN}, OverFeat~\cite{Overfeat}, PNASNet-5 (4, 216)~\cite{PNASNet}, PolyNet~\cite{PolyNet}, PreResNet-152~\cite{preresnet}, PreResNet-200~\cite{preresnet}, PyramidNet-101 (alpha=250)~\cite{PyramidNet}, PyramidNet-200 (alpha=300)~\cite{PyramidNet}, PyramidNet-200 (alpha=450)~\cite{PyramidNet}, ResNet-152~\cite{ResNet}, ResNet-50~\cite{ResNet}, ResNet-101~\cite{ResNet}, ResNeXt-101, SENet~\cite{SENet}, ShuffleNet (1.5)~\cite{ShuffleNet}, ShuffleNet (x2)~\cite{ShuffleNet}, SimpleNet~\cite{SimpleNet}, SqueezeNet~\cite{iandola2016squeezenet}, SqueezeNetv1.1~\cite{iandola2016squeezenet}, SqueezeNext (1.0-23v5)~\cite{SqueezeNext}, SqueezeNext (2.0-23)~\cite{SqueezeNext}, SqueezeNext (2.0-23v5)~\cite{SqueezeNext}, TinyDarkNet~\cite{TinyDarkNet}, VGG16~\cite{VGG16}, Xception~\cite{Xception}, ZynqNet~\cite{ZynqNet}.

In this study, the units used for $p(\mathcal{N})$ and $m(\mathcal{N})$ for two of the quantitative performance metrics (information density and the proposed NetScore metric) are in M-Params (millions of parameters) and G-MACs (billions of MAC operations), respectively, given that most modern deep convolutional neural networks are within those architectural and computational complexity ranges.

\begin{figure}[!tp]
	\begin{center}
		\includegraphics[width = 15 cm]{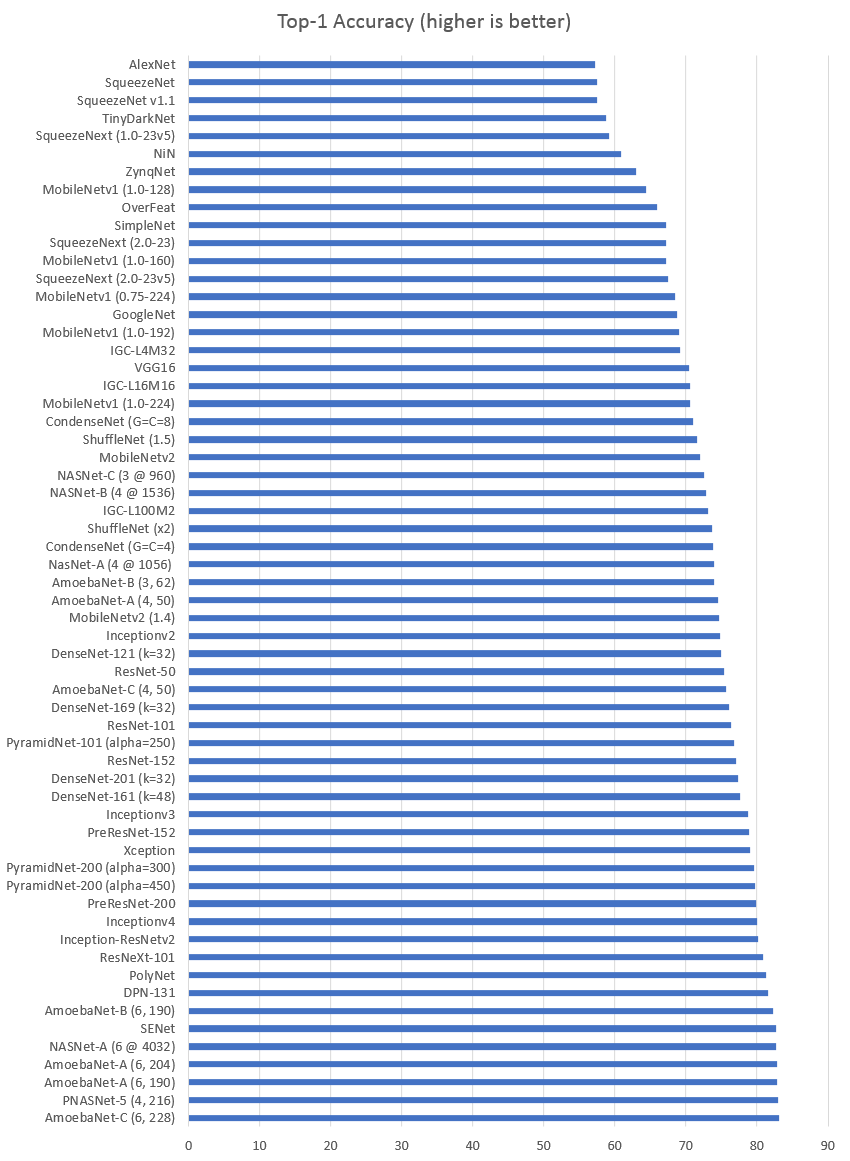}
	\end{center}
	\caption{Top-1 accuracy across 60 different deep convolutional neural networks for the ILSVRC 2012 dataset.}
	\label{fig:top1}
\end{figure}

\textbf{Top-1 accuracy}:  The top-1 accuracies across 60 different deep convolutional neural networks for the ILSVRC 2012 dataset is shown in Fig.~\ref{fig:top1}.  It can be clearly observed that significant progress has been made in the design of deep convolutional neural networks for image classification over the past six years, with the difference between the deep convolutional neural network with the highest top-1 accuracy in this study (i.e., AmoebaNet-C (6, 228)) and that of AlexNet exceeding 25\%.  It is also interesting to see that more recent developments in efficient deep convolutional neural networks such as MobileNetv1, MobileNetv2, and ShuffleNet all have top-1 accuracies that exceed VGG-16, the third largest tested network evaluated in the study that was also the state-of-the-art just four years ago, thus further illustrating the improvements in network design over the past few years.

\begin{figure}[!tp]
	\begin{center}
		\includegraphics[width = 15 cm]{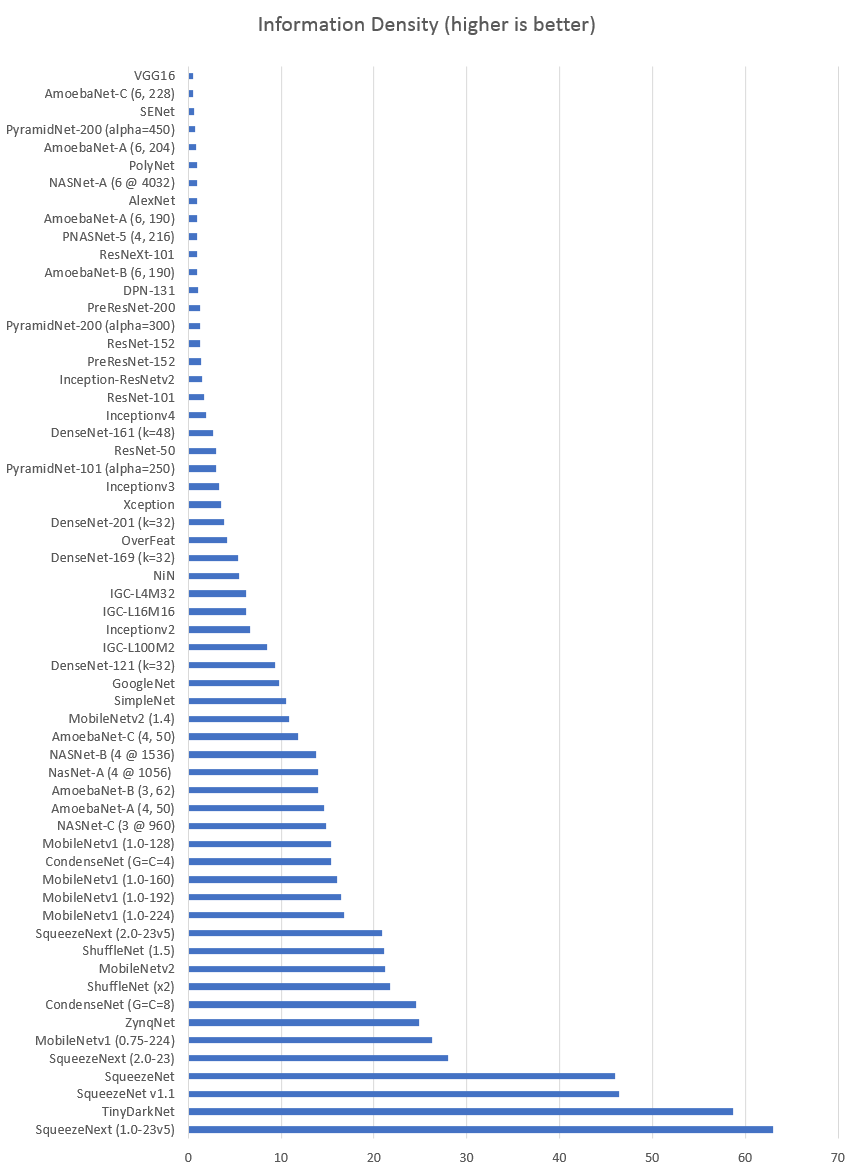}
	\end{center}
	\caption{Information density across 60 different deep convolutional neural networks for the ILSVRC 2012 dataset. Units are in \%/M-Params.}
	\label{fig:infodensity}
\end{figure}

\textbf{Information density}: The information densities across 60 different deep convolutional neural networks for the ILSVRC 2012 dataset is shown in Fig.~\ref{fig:infodensity}.  It can be clearly observed that the deep convolutional neural networks that were specifically designed for efficiency (e.g., MobileNetv1, MobileNetv2, ShuffleNet, SqueezeNet, Tiny DarkNet, and SqueezeNext) have significantly higher information densities compared to networks that were designed purely with accuracy as a metric.  More specifically, the SqueezeNext (1.0-23v5), Tiny DarkNet, and the SqueezeNet family of networks had the highest information density by a wide margin compared to the other tested deep convolutional neural networks, which can be attributed to their significantly lower architectural complexity in terms of number of network parameters.  Another notable observation from the results in Fig.~\ref{fig:infodensity} is that the dynamic range of the information density metric is quite large across the diverse set of 60 deep convolutional neural networks evaluated in this study.

\begin{figure}[!tp]
	\begin{center}
		\includegraphics[width = 15 cm]{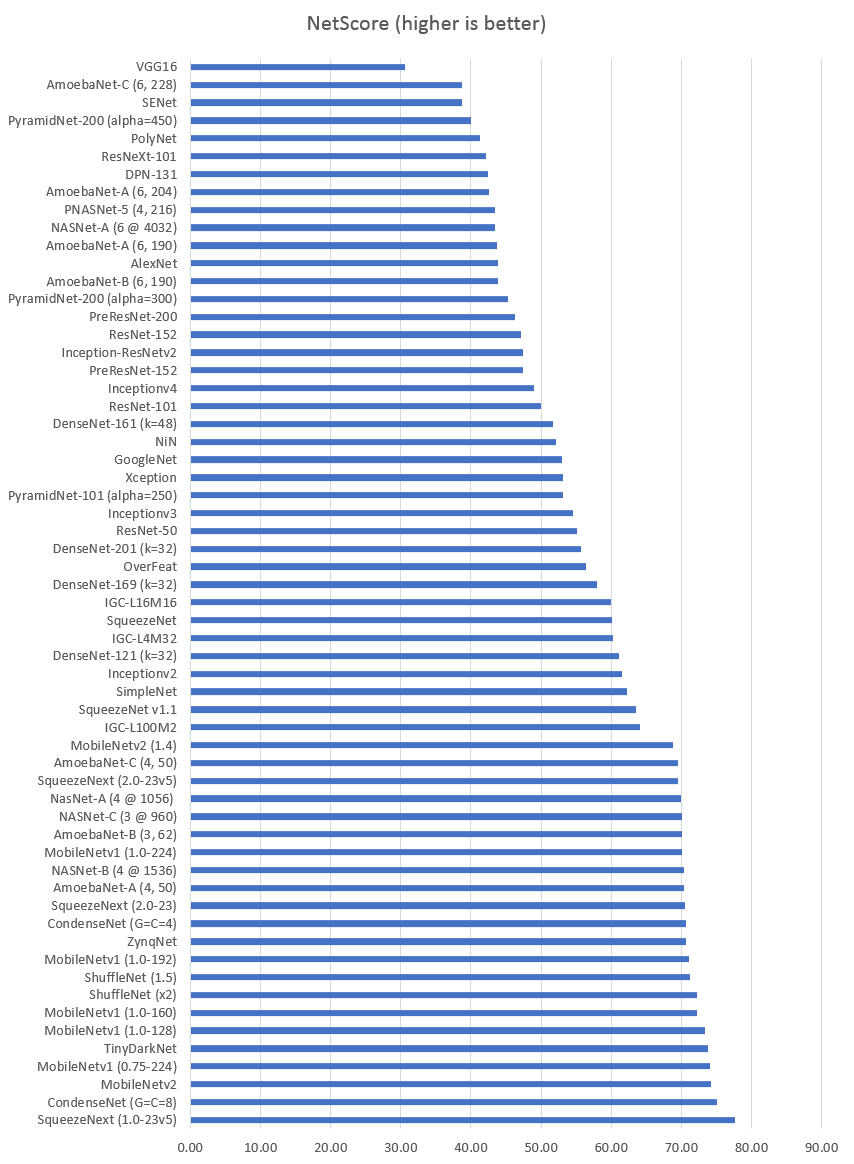}
	\end{center}
	\caption{NetScore across 60 different deep convolutional neural networks for the ILSVRC 2012 dataset.}
	\label{fig:Netscore}
\end{figure}

\textbf{NetScore}: The NetScore across 60 different deep convolutional neural networks for the ILSVRC 2012 dataset is shown in Fig.~\ref{fig:Netscore}.  Similar to the trend observed in Fig.~\ref{fig:infodensity}, it can be clearly observed that many of the deep convolutional neural networks that were specifically designed for efficiency have significantly higher NetScores compared to networks that were designed purely with accuracy as a metric.  However, what is interesting to observe is that the NetScore ranking amongst these efficient networks are quite different than that when using the information density metric.  In particular, the top ranking deep convolutional neural networks with the highest NetScores are SqueezeNext (1.0-23v5), CondenseNet (G=C=8), and MobileNetv2.

The SqueezeNet family of networks, on the other hand, had much lower relative NetScores compared to the aforementioned efficient networks despite having the top two highest information densities.  This observation illustrates the effect of incorporating computational complexity to the assessment of deep convolutional neural networks for practical usage, given that while the SqueezeNet family of networks has significantly lower architectural complexities compared to other tested networks, it also is offset by noticeably higher computational complexities compared to other tested efficient networks such as the MobileNetv1, MobileNetv2, SqueezeNext, and ShuffleNet network families.

The proposed NetScore metric, which by no means is perfect, could potentially be useful for guiding practitioners in model search and design and hopefully push the conversation towards better universal metrics for evaluating deep neural networks for use in practical scenarios.  Future work includes incorporating additional or alternative factors that are important to assessing architectural and computational complexities of deep neural networks beyond what is being used in the NetScore metric, as well as finding a good balance between these different factors based on relative importance for the deployment of deep neural networks for practical usage in scenarios such as mobile devices and other edge devices.

\vspace{-0.35 cm}
\section*{Acknowledgment}
\vspace{-0.15 cm}
The author thank Natural Sciences and Engineering Research Council of Canada (NSERC), the Canada Research Chairs program, Nvidia, and DarwinAI.

{\small
\bibliographystyle{plain}
\bibliography{ccn_style}

\begin{thebibliography}{10}

\bibitem{Canziani}
A.~Canziani, A.~Paszke, and E.~Culurciello.
\newblock An analysis of deep neural network models for practical applications.
\newblock {\em arXiv preprint arXiv:1605.07678}, 2017.

\bibitem{DPN}
Yunpeng Chen, Jianan Li, Huaxin Xiao, Xiaojie Jin, Shuicheng Yan, and Jiashi
  Feng.
\newblock Dual path networks.
\newblock {\em CoRR}, abs/1707.01629, 2017.

\bibitem{Xception}
Fran{\c{c}}ois Chollet.
\newblock Xception: Deep learning with depthwise separable convolutions.
\newblock {\em CoRR}, abs/1610.02357, 2016.

\bibitem{lecun2015deep}
Y.~Le Cun, Y.~Bengio, and G.~Hinton.
\newblock Deep learning.
\newblock {\em Nature}, 2015.

\bibitem{lecun2}
Y.~Le Cun, L.~Bottou, Y.~Bengio, and P.~Haffner.
\newblock Gradient-based learning applied to document recognition.
\newblock {\em Proceedings of IEEE}, 1998.

\bibitem{lecun1}
Y.~Le Cun, J.~Denker, D.~Henderson, R.~Howard, W.~Hubbard, and L.~Jackel.
\newblock Handwritten digit recognition with a back-propagation network.
\newblock {\em Proceedings of the Advances in Neural Information Processing
  Systems (NIPS)}, 1989.

\bibitem{SqueezeNext}
Amir Gholami, Kiseok Kwon, Bichen Wu, Zizheng Tai, Xiangyu Yue, Peter~H. Jin,
  Sicheng Zhao, and Kurt Keutzer.
\newblock Squeezenext: Hardware-aware neural network design.
\newblock {\em CoRR}, abs/1803.10615, 2018.

\bibitem{ZynqNet}
D.~Gschwend.
\newblock Zynqnet: An fpga-accelerated embedded convolutional neural network.
\newblock {\em https://github.com/dgschwend/zynqnet}, 2016.

\bibitem{PyramidNet}
Dongyoon Han, Jiwhan Kim, and Junmo Kim.
\newblock Deep pyramidal residual networks.
\newblock {\em CoRR}, abs/1610.02915, 2016.

\bibitem{SimpleNet}
Seyyed~Hossein HasanPour, Mohammad Rouhani, Mohsen Fayyaz, and Mohammad
  Sabokrou.
\newblock Lets keep it simple, using simple architectures to outperform deeper
  and more complex architectures.
\newblock {\em CoRR}, abs/1608.06037, 2016.

\bibitem{MaskRCNN}
K.~He, G.~Gkioxari, P.~Dollar, and R.~Girshick.
\newblock Mask r-cnn.
\newblock {\em {ICCV}}, 2017.

\bibitem{ResNet}
Kaiming He, Xiangyu Zhang, Shaoqing Ren, and Jian Sun.
\newblock Deep residual learning for image recognition.
\newblock {\em CoRR}, abs/1512.03385, 2015.

\bibitem{preresnet}
Kaiming He, Xiangyu Zhang, Shaoqing Ren, and Jian Sun.
\newblock Identity mappings in deep residual networks.
\newblock {\em CoRR}, abs/1603.05027, 2016.

\bibitem{howard2017mobilenets}
A.~Howard, M.~Zhu, B.~Chen, D.~Kalenichenko, W., T.~Weyand, M.~Andreetto, and
  H.~Adam.
\newblock Mobilenets: Efficient convolutional neural networks for mobile vision
  applications.
\newblock {\em arXiv preprint arXiv:1704.04861}, 2017.

\bibitem{SENet}
Jie Hu, Li~Shen, and Gang Sun.
\newblock Squeeze-and-excitation networks.
\newblock {\em CoRR}, abs/1709.01507, 2017.

\bibitem{CondenseNet}
Gao Huang, Shichen Liu, Laurens van~der Maaten, and Kilian~Q. Weinberger.
\newblock Condensenet: An efficient densenet using learned group convolutions.
\newblock {\em CoRR}, abs/1711.09224, 2017.

\bibitem{densenet}
Gao Huang, Zhuang Liu, and Kilian~Q. Weinberger.
\newblock Densely connected convolutional networks.
\newblock {\em CoRR}, abs/1608.06993, 2016.

\bibitem{iandola2016squeezenet}
F.~Iandola, S.~Han, M.~Moskewicz, K.~Ashraf, W.~Dally, and K.~Keutzer.
\newblock Squeezenet: Alexnet-level accuracy with 50x fewer parameters and< 0.5
  mb model size.
\newblock {\em arXiv preprint arXiv:1602.07360}, 2016.

\bibitem{krizhevsky2012imagenet}
A.~Krizhevsky, I.~Sutskever, and G.~Hinton.
\newblock Imagenet classification with deep convolutional neural networks.
\newblock In {\em {NIPS}}, 2012.

\bibitem{NiN}
Min Lin, Qiang Chen, and Shuicheng Yan.
\newblock Network in network.
\newblock {\em CoRR}, abs/1312.4400, 2013.

\bibitem{PNASNet}
Chenxi Liu, Barret Zoph, Jonathon Shlens, Wei Hua, Li{-}Jia Li, Li~Fei{-}Fei,
  Alan~L. Yuille, Jonathan Huang, and Kevin Murphy.
\newblock Progressive neural architecture search.
\newblock {\em CoRR}, abs/1712.00559, 2017.

\bibitem{liu2016ssd}
W.~Liu, D.~Anguelov, D.~Erhan, C.~Szegedy, S.~Reed, C.~Fu, and A.~Berg.
\newblock {SSD}: Single shot multibox detector.
\newblock In {\em {ECCV}}, 2016.

\bibitem{ImageNet}
H.~Su J. Krause S. Satheesh S. Ma Z. Huang A. Karpathy A. Khosla M. Bernstein
  et al. journal={International Journal of Computer Vision}~year={2015}
  O.~Russakovsky, J.~Deng.
\newblock Imagenet large scale visual recognition challenge.

\bibitem{AmoebaNet}
Esteban Real, Alok Aggarwal, Yanping Huang, and Quoc~V. Le.
\newblock Regularized evolution for image classifier architecture search.
\newblock {\em CoRR}, abs/1802.01548, 2018.

\bibitem{TinyDarkNet}
J.~Redmon.
\newblock Tiny darknet.
\newblock {\em https://pjreddie.com/darknet/tiny-darknet/}, 2016.

\bibitem{sandler2018mobilenetv2}
M.~Sandler, A.~Howard, M.~Zhu, A.~Zhmoginov, and L.~Chen.
\newblock Mobilenetv2: Inverted residuals and linear bottlenecks.
\newblock {\em arXiv preprint arXiv:1704.04861}, 2017.

\bibitem{Overfeat}
Pierre Sermanet, David Eigen, Xiang Zhang, Micha{\"{e}}l Mathieu, Rob Fergus,
  and Yann LeCun.
\newblock Overfeat: Integrated recognition, localization and detection using
  convolutional networks.
\newblock {\em CoRR}, abs/1312.6229, 2013.

\bibitem{squishednet}
M.~Shafiee, F.~Li, B.~Chwyl, and A.~Wong.
\newblock Squishednets: Squishing squeezenet further for edge device scenarios
  via deep evolutionary synthesis.
\newblock In {\em {NIPS}}, 2017.

\bibitem{VGG16}
Karen Simonyan and Andrew Zisserman.
\newblock Very deep convolutional networks for large-scale image recognition.
\newblock {\em CoRR}, abs/1409.1556, 2014.

\bibitem{Inceptionv4}
Christian Szegedy, Sergey Ioffe, and Vincent Vanhoucke.
\newblock Inception-v4, inception-resnet and the impact of residual connections
  on learning.
\newblock {\em CoRR}, abs/1602.07261, 2016.

\bibitem{GoogleNet}
Christian Szegedy, Wei Liu, Yangqing Jia, Pierre Sermanet, Scott~E. Reed,
  Dragomir Anguelov, Dumitru Erhan, Vincent Vanhoucke, and Andrew Rabinovich.
\newblock Going deeper with convolutions.
\newblock {\em CoRR}, abs/1409.4842, 2014.

\bibitem{Inceptionv2}
Christian Szegedy, Vincent Vanhoucke, Sergey Ioffe, Jonathon Shlens, and
  Zbigniew Wojna.
\newblock Rethinking the inception architecture for computer vision.
\newblock {\em CoRR}, abs/1512.00567, 2015.

\bibitem{munet}
Alexander Wong, Mohammad~Javad Shafiee, and Michael~St. Jules.
\newblock {muNet}: {A} highly compact deep convolutional neural network
  architecture for real-time embedded traffic sign classification.
\newblock {\em CoRR}, abs/1804.00497, 2018.

\bibitem{tinyssd}
Alexander Wong, Mohammad~Javad Shafiee, Francis Li, and Brendan Chwyl.
\newblock Tiny {SSD:} {A} tiny single-shot detection deep convolutional neural
  network for real-time embedded object detection.
\newblock {\em CoRR}, abs/1802.06488, 2018.

\bibitem{IGC}
Ting Zhang, Guo{-}Jun Qi, Bin Xiao, and Jingdong Wang.
\newblock Interleaved group convolutions for deep neural networks.
\newblock {\em CoRR}, abs/1707.02725, 2017.

\bibitem{ShuffleNet}
Xiangyu Zhang, Xinyu Zhou, Mengxiao Lin, and Jian Sun.
\newblock Shufflenet: An extremely efficient convolutional neural network for
  mobile devices.
\newblock {\em CoRR}, abs/1707.01083, 2017.

\bibitem{PolyNet}
Xingcheng Zhang, Zhizhong Li, Chen~Change Loy, and Dahua Lin.
\newblock Polynet: {A} pursuit of structural diversity in very deep networks.
\newblock {\em CoRR}, abs/1611.05725, 2016.

\bibitem{NASNet}
Barret Zoph, Vijay Vasudevan, Jonathon Shlens, and Quoc~V. Le.
\newblock Learning transferable architectures for scalable image recognition.
\newblock {\em CoRR}, abs/1707.07012, 2017.

\end{thebibliography}
}

\end{document}